\documentclass[review]{elsarticle}

\usepackage{lineno,hyperref}
\usepackage{mathrsfs}
\usepackage{algorithm}  
\usepackage{algpseudocode}  
\usepackage{amsmath}  
\usepackage{amsfonts}

\usepackage{pgfplots}
\usepackage{caption}
\usepackage{subcaption}
\usepackage{microtype}
\usepackage{multirow}

\journal{Journal of \LaTeX\ Templates}

\begin{document}

\begin{frontmatter}

\title{Progressive Sample Mining and Representation Learning for One-Shot Person Re-identification with Adversarial Samples}

\author[mymainaddress]{Hui Li}
\ead{hui.li02@xjtlu.edu.cn}

\author[mymainaddress]{Jimin Xiao\corref{mycorrespondingauthor}}
\ead{jimin.xiao@xjtlu.edu.com}

\author[mymainaddress]{Mingjie Sun}

\author[mymainaddress]{Eng Gee Lim}

\author[mysecondaryaddress]{Yao Zhao}
\cortext[mycorrespondingauthor]{Corresponding author}

\address[mymainaddress]{Xi'an Jiaotong-Liverpool University,
Suzhou,
China}
\address[mysecondaryaddress]{Beijing Jiaotong University,
Beijing, China}

\begin{abstract}
In this paper, we aim to tackle the one-shot person re-identification problem where only one image is labelled for each person, while other images are unlabelled. This task is  challenging due to lack of sufficient labelled training data. To tackle this problem, we propose to iteratively guess \emph{pseudo} labels for the \emph{unlabeled} image samples, which are later used to update the re-identification model together with the \emph{labeled} samples. A new sampling mechanism is designed to select \emph{unlabeled} samples to \emph{pseudo labeled} samples based on the distance matrix, and to form a training triplet batch including both \emph{labeled} samples and \emph{pseudo} labelled samples. 
We also design an HSoften-Triplet-Loss to soften the negative impact of the incorrect \emph{pseudo} label, considering the unreliable nature of \emph{pseudo labeled} samples. 
Finally, we deploy an adversarial learning method to expand the image samples to different camera views. 
Our experiments show that our framework achieves a new state-of-the-art one-shot Re-ID performance on Market-1501 (mAP 42.7\%) and DukeMTMC-Reid dataset (mAP 40.3\%). Code will be available soon.
\end{abstract}


\end{frontmatter}


\section{Introduction}
Person re-identification (Re-ID) has attracted increasing attentions from both academia and industry due to its essential applications on public security and surveillance. Along with the wide deployment of visual surveillance, Re-ID becomes one of the key research topics in the computer vision community.

The Re-ID task aims to match people with variations of cameras, scales and views. Some recent methods \cite{Xu2018AttentionAwareCN, Wu2018DeepAF, Song_2018,lin2019improving, ding2019feature} have been proved effective in learning a robust feature representation to distinguish the high similar appearance of different people. 
However, training Re-ID model with acceptable accuracy using fully supervised learning requests dozens of training samples for each class/identity \cite{LeCun2015}.

Collecting large amount of training data is neither cheap nor reliable using human annotations. Labels with ambiguity and inconsistency may be annotated due to lack of annotating experience. At the same time, manual annotation may request to collect sensitive and privacy information, such as pedestrian portrait, location or identity. Such privacy information is in the risk of exposure to the public. Therefore, researchers begin to study semi-supervised learning methods to use the samples in a more efficient way.

In recent years, the research focus of Re-ID changes from fully-supervised learning to domain adaptation \cite{Fan_2018, deng_2018, wang_2018, zhong2019invariance, fu2018selfsimilarity}, weakly-supervised learning \cite{meng2019weakly} and one-shot learning \cite{wu2019progressive}.
Domain adaptation methods try to reduce the discrepancy between the source domain and the target domain. For example, Fu et al. \cite{fu2018selfsimilarity} proposed a self-similarity grouping process on global and local parts from both the source and target domains.  Deng et al. \cite{deng_2018} chose GAN to translate the source domain images to target domain and train the generated images with the original labels. Wang et al. \cite{wang_2018} used the classification score of a model trained on auxiliary data to form pair-wises for the unlabeled persons. Meng et al. \cite{meng2019weakly} considered weakly supervised person Re-ID modeling, where we only know that an identity appears in a video without the requirement of annotating the identity in any frame of the video during the training procedure. 

\begin{figure}[H]
  \centering
  \includegraphics[width=1\linewidth,trim=400 800 370 700,clip]{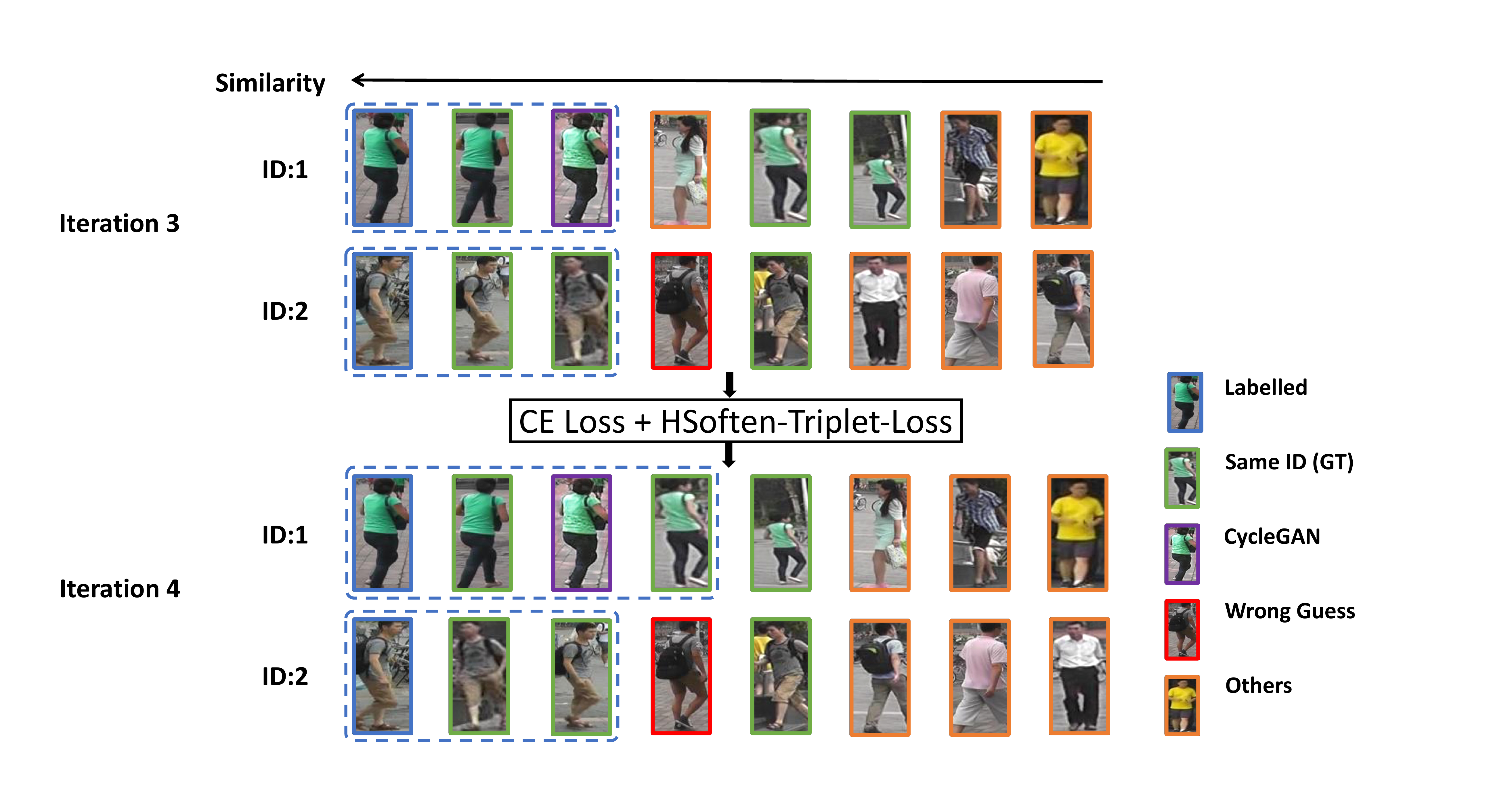}
   \caption{Example of the selection process for \emph{pseudo labelled} person.  
   The upper part is the third iteration step, where we choose 2 similar images with the same \emph{pseudo} ID. After one more training iteration, in the lower part, we aim to choose one more image with \emph{pseudo} label for each person, but ignore the wrong sample for ID 2.
   }
\label{fig:long}
\end{figure}

One-shot Re-ID learning requires the minimum label annotation. Video based one-shot person re-identification has witnessed a significant progress  \cite{8237528,8611378, chen2018deep, inproceedings, zhao2019similarity, wu2018cvpr_oneshot}, with
\cite{wu2018cvpr_oneshot} achieving a high score close to the supervised baseline.  However, for image based Re-ID task \cite{wu2019progressive}, it is impossible to form a tracklet with the same ID.
Only one sample is annotated with ground-truth label for each class in an image based one-shot Re-ID task, while other samples are unlabeled \cite{wu2019progressive}. During the iteratively training process, some unlabeled samples are selected and annotated with \emph{pseudo} label. Re-id models will be updated using samples with both ground-truth and \emph{pseudo} labels. However, the obtained results of \cite{wu2019progressive}  are still far away from fully-supervised learning, mainly due to the three challenges for this task: 1) how to select unlabeled samples for \emph{pseudo} label; 2) how to design loss functions for  semi-supervised training; 3) how to overcome the overfitting problem due to lack of data. 

To tackle the above challenges, in this paper, we propose a new framework for the image based one-shot Re-ID task, the example of our framework is shown on Fig.~\ref{fig:long}. More specifically, we first design an iterative selection strategy to expend \emph{unlabelled} sample to the \emph{pseudo labeled} samples, from very confident samples to all \emph{unlabeled} samples, followed by a model update using the new \emph{pseudo} samples. Also, specifically designed for semi-supervised learning, we introduce an HSoften-Triplet-Loss to mitigate the negative effects of incorrect guessed \emph{pseudo} labels. Finally, we use adversarial learning to generate more views of different cameras to enrich the training data and avoid overfitting. Though \emph{pseudo} label generation exists in other semi-supervised methods \cite{wu2019progressive}, our method is specifically designed for one-shot person re-identification task and achieves the state-of-the-art performance. 

Our contributions are four-fold:
\begin{itemize}
\item We propose a \emph{pseudo} label sampling framework for one-shot Re-ID, which is based on the relative sample distance to the feature center of each labelled class, by taking the adversarial samples into consideration.

\item To the best of our knowledge, our work is the first work to deploy the data augmentation of the adversarial learning on image-based one-shot person Re-ID task. Our experiment proves its effectiveness.

\item Considering the nature of \emph{pseudo} labels, we introduce an HSoften-Triplet-Loss to soften the negative influence of incorrect \emph{pseudo} label. Meanwhile, a new batch formation rule is designed by taking different nature of \emph{labelled} samples and \emph{pseudo labelled} samples into account. 

\item We achieve the state-of-the-art mAP score of 42.7\% on Market1501 and 40.3\% on DukeMTMC-Reid, 16.5 and 11.8 higher than EUG \cite{wu2019progressive} respectively. 

\end{itemize}

\section{Related Work}
\subsection{Person Re-ID }
Recently, following the success of Convolutional Neural Network, many approaches based on deep learning methods proved the effective on the person Re-ID task, both for fully-supervised learning and semi-supervised learning. For \emph{fully-supervised} learning, many new methods have made great improvement to reach a better performance like \cite{su2017pose,liu2018pose,Kalayeh_2018, song2018mask, Lin_2019, zhang2018person}. Person key-points are exploited for person Re-ID, where Su et al. \cite{su2017pose} used the human positions to generate more images from different views, and Liu et al. \cite{liu2018pose} separated the human body into six main parts based on the key-points. \cite{Kalayeh_2018} and \cite{song2018mask} are mask-based method, where border information is used rather than the bounding box to remove the background noise. There are also some attribute-based methods like \cite{Lin_2019} and \cite{zhang2018person}, where semantic descriptions of a person are exploited,  from global attributes, like gender or age, to a more identical multi-level attributes, like the color of shoes. 

In terms of \emph{semi-supervised} Re-ID methods, recent researches interest in transfer learning \cite{yu2019unsupervised, song2018unsupervised, Li_2019}, which have labels on the source domain but have no labels for the target domain. Yu et al. \cite{yu2019unsupervised} trained the model from source domain, and predicted both the feature and classification score on the target domain to find the relations within the target domain. Song et al. \cite{song2018unsupervised} tried to find the cluster of the feature from both source and target domain to group the same person together. Li et al. \cite{Li_2019} found another way to combine the problem of the person Re-ID task with the tracking task, and used the pre-trained tracking model with the source domain to group the people in the target domain. 

The design of loss functions also influences the person Re-ID task. Three most common losses for person Re-ID task are softmax cross-entropy loss, centre loss \cite{10.1007/978-3-319-46478-7_31} and triplet loss \cite{alex2017defense}. \emph{Softmax loss} initially comes from the classification task and is the basic loss to train the task with identity information. Though the person Re-ID task is similar to classification task because they both have identities, for the person Re-ID task, the IDs between training and inference are totally different, which makes the softmax loss not perfectly suitable for the person Re-ID task. So, Wen et al. \cite{10.1007/978-3-319-46478-7_31} introduced the centre loss, which had a better centre clustering property to solve the problem. Later, it has been proved that triplet loss has a better feature presentation learning ability, not only to pull image features of the same ID together but also to push the different people away \cite{alex2017defense}. Thus, the basic loss for person Re-ID task is the combination of both softmax loss and triplet loss. 

Different from previous works, our work designs a new training data sampling and generating mechanism, and a loss function specifically designed for the one-shot person Re-ID task, by taking the nature of various samples into consideration. 

\subsection{GAN and Re-ID }
The Generative Adversarial Nets (GAN) \cite{goodfellow2014generative} has been adopted in many applications including image generation \cite{brock2018large,8578190}, image-to-image translation \cite{Isola_2017, liu2016coupled}, style transfer \cite{Gatys_2016, Dong_2018, CycleGAN2017} and so on. Apart from generating fake images, researchers found that GAN is also useful to achieve a universal improvement on many tasks like super-resolution \cite{Ledig_2017}, semantic segmentation \cite{luc2016semantic} and so on.

For the person Re-ID task, the main applications of GAN are camera-based person image augmentation and pose-guided person image augmentation \cite{ Zhong_2018, ge2018fdgan}. With the great success of CycleGAN \cite{CycleGAN2017}, which can transfer an image style to another one without complex feature engineering, \cite{ Zhong_2018} generated more person images of different cameras to enrich the dataset. On the other hand, because of the limited number of images for each person, it is pretty hard to obtain many images of a person under different camera views.  \cite{ge2018fdgan} proposed a pose-guided GAN to generate images of different poses for each person.

Compared with the previous methods that only consider generating new images as data augmentation, we integrate the newly generated image to our sampling process specifically designed for the one-shot Re-ID task.

\section{Method}
\begin{figure}[H]
\begin{center}
  \includegraphics[width=1.1\linewidth,trim=400 800 450 700,clip,angle=270]{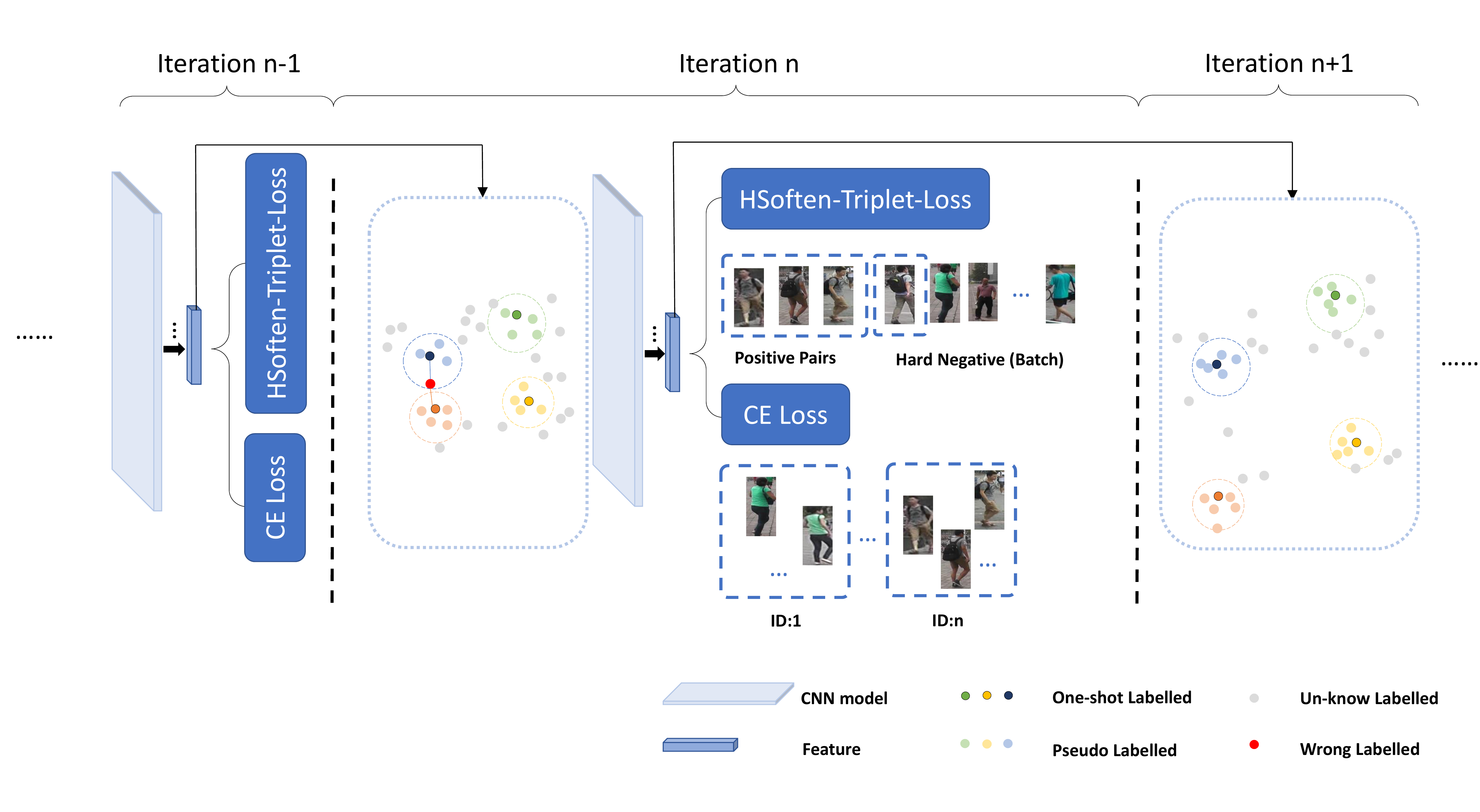}
\end{center}
   \caption{Overview of our method. Our training process takes several iterations. Each iteration has two main steps: 1) Add  \emph{pseudo labelled} images for each \emph{labeled} image. 2) Train the model with both CE loss and HSoft-triplet loss. After each iteration, the model should be more discriminative for feature representation and more reliable to generate the next similarity matrix. This is demonstrated by the fact that image features of the same person are clustered in a more compact manner, and features of different person move apart.  
   The new similarity matrix is used to sample more \emph{pseudo labelled} images for the next iteration training. Best viewed in color.
   }
\label{fig:framework}
\end{figure}

The overview of our framework is shown in Fig.~\ref{fig:framework}. Our algorithm is described in Algorithm.~\ref{algorithm_all}. There are two main steps for each iteration of the training process. Firstly, the similarity matrix is evaluated, based on the model obtained from the previous iteration. Based on the similarity matrix, several samples are selected from the \emph{unlabelled} set, and \emph{pseudo} labels are assigned. Secondly, a new model is trained with samples including the latest added samples with \emph{pseudo} label. 

Technically, our framework proposes a new sample mining process specifically designed for one-shot Re-ID (\ref{process}).Meanwhile, an adversarial training mechanism using camera ID is proposed to generate more views of different cameras to enrich the training data, and to avoid the overfitting problem (\ref{GAN}). A Soften-Triplet-Loss is also specifically designed for the semi-supervised person Re-ID task (\ref{loss}). 

\begin{algorithm}[H]
  \caption{Overall Algorithm}  
  \label{alg:Training}  
 \begin{algorithmic}  
    \Require  
    The dataset with both \emph{labelled} (one-shot) and \emph{unlabelled} samples.
    \Ensure  
    The feature representation model  $\phi(\theta;x)$ 
    \State \textbf{pre-initial}
    \State Training the adversarial model and generate more data for different cameras.
    \State \textbf{begin}
    \State Train the initial model using the one-shot labelled samples  
    \Repeat 
      \State Load the latest model
      \State Calculate the distance matrix
      \State Do sample mining based on the distance matrix
      \State Reload the ImageNet pre-trained model 
    \Repeat  
      \For {Sample the batch randomly from \emph{labelled} samples}
        \State Form triplets for labelled samples based on the similarity matrix.
        \State Calculate the Softmax loss
        \State Calculate the HSoften-Triplet-Loss
        \State Update the model by minimising losses
      \EndFor
    \Until{Model well trained for this selection iteration}
    \Until{All unlabelled images get pseudo labels}
  \end{algorithmic}  
\label{algorithm_all}
\end{algorithm}

\subsection{Vanilla Pseudo Label Sampling (PLS)}
\label{process}

\subsubsection{Problem Definition} 

We define the whole training dataset as $\mathcal{X}$ where $x$ ($x \in \mathcal{X}$) is one image from the dataset, and $N$ is the number of training images. The whole dataset can be divided into two parts, including the \emph{labelled} part as $\mathcal{L}$ ($l \in \mathcal{L}$)  and
the remained \emph{unlabelled} part as $ \mathcal{U}$ ($u \in \mathcal{U}$). In the \emph{labelled} part $\mathcal{L}$, we have $C$ classes, each of which has one labeled image.
To evaluate the distance between $\mathcal{L}$ and $\mathcal{U}$, we also define the distance matrix $M \in \mathbb{R}^{C \times{(N-C)}}$

\begin{equation}
\begin{split}
M\mathbf{}\mathit{} = \left [ \begin{array}{cccc}\
M[0,0] & M[0,1] & ... & M[0,N-C]\\
M[1,0] & M[1,1] & ... & M[1,N-C]\\
... & ... & ... & ...\\
M[C,0] & M[C,0] & ... & M[C,N-C]\end{array}
\right ]
\label{eq:distance}, 
\end{split}
\end{equation} 
where $M[i,j]$ denotes the L2 feature distance between the labeled sample $l_i$ ($l_i \in \mathcal{L}$) and the unlabeled sample $u_j$ ($u_j \in \mathcal{U} $). 

The objective of our Re-ID task is to train a feature representation model $\phi(\theta;x)$ using $ \mathcal{L}$ and  $ \mathcal{U}$. 
To train this model, we also define the classifier model $f(W;\phi (\theta;x))$, where $W$ is the parameter set for the classifier.

\subsubsection{Sample Mining}

In the initial model training step, we only use the \emph{labelled} images 
$\mathcal{L}$ to train $\phi(\theta;x)$  and $f(W;\phi (\theta;x))$. 
By doing this, we expect to obtain higher accuracy for the following guesses. 
Training the model using samples with incorrect predicted labels undermine the performance dramatically, as it will influence the following pseudo label sampling process.

In the following step, we iteratively sample pseudo labels, and train our model with new samples. For iteration $T$, we select $TC$ unlabelled samples from $\mathcal{U}$ to generate $\mathcal{U}^p$, and assign them preliminary pseudo labels. For each labeled sample $l$ ($l \in \mathcal{L}$), we select $T$ closest samples from $\mathcal{U}$ based on distance matrix $M$.

We also use a relative distance to refine the samples with preliminary pseudo labels $\mathcal{U}^p$. The samples with pseudo labels we choose for each person should be close to its reference, yet far away from other references. 
\begin{equation}
\begin{split}
\mathcal{U}^{sp}=\{ u^p_{i_j} | u^p_{i_j} \in \mathcal{U}^p; M[i,j]< M[k,j] \\
(\forall k \in C \& k \neq i) \} 
\label{eq:refine},
\end{split}
\end{equation}
where $u^p_{i_j}$ is a sample with the preliminary pseudo label of class $i \in C $, and with sample index $j$. After the sample mining, we train model  $\phi(\theta;x)$ and \\ $f(W;\phi (\theta;x))$ using labeled images  $\mathcal{L}$ together with our selected pseudo images $\mathcal{U}^{sp}$. We update our distance matrix using the newly updated model $\phi(\theta;x)$, and start the next iteration.

\subsection{PLS with Adversarial Learning}
\label{GAN}

In our framework, we also apply the adversarial learning into the one-shot Re-ID task. To be more specific, we use the CycleGAN \cite{CycleGAN2017} as data augmentation tool
to generate images of different cameras, and adapt the enhanced dataset to our PLS framework.

\subsubsection{Adversarial Generated Samples} 
Our main purpose of adversarial learning is to enrich the original dataset $\mathcal{X}$  to $\mathcal{\hat{X}}$ before training $\phi(\theta;x)$. The dataset $\mathcal{X}$ can be described as $\mathcal{X}= \left \{\mathcal{X}^{cam}_{a} \right \}^A_{a=1}$ where $A$ is the number of cameras. For $\mathcal{X}^{cam}_{a}$ ($x^{cam}_a \in \mathcal{X}^{cam}_{a}$), we want to use $x^{cam}_a$ from camera $a$ to generate fake images for all other $(A-1)$ cameras. In total, it will take $\mathbb{C}_{A}^{2}$ training pairs among all $A$ cameras.
In fact, the camera ID used for our adversarial learning is very easy to obtain, which does not require human annotation.  

In CycleGAN \cite{CycleGAN2017}, when considering training the adversarial network between cameras $a$ and $\hat{a}$, we need two generator networks $G:a \rightarrow \hat{a}$ and $F:\hat{a}\rightarrow a$, using the discriminators $D_{G}$ and $D_{F}$ respectively. The total CycleGAN loss will be:
\begin{equation}
\begin{split}
L(G,F,D_{G},D_{F},\mathcal{X}^{cam}_{a},\mathcal{X}^{cam}_{\hat{a}}) = \\ L_{\emph{GAN}}(D_{G},G,\mathcal{X}^{cam}_{a},\mathcal{X}^{cam}_{\hat{a}})\\
+ L_{\emph{GAN}}(D_{F},F,\mathcal{X}^{cam}_{\hat{a}},\mathcal{X}^{cam}_{a}) \\
+ \lambda L_{cyc}(G,F)
\end{split}
\label{eq:CycGAN}
\end{equation} 
where $L_{\emph{GAN}}(D_{G},G,\mathcal{X}^{cam}_{a},\mathcal{X}^{cam}_{\hat{a}})$ is the generative adversarial loss between the discriminators $D_{G}$ and the generator $G$. $L_{\emph{GAN}}(D_{F},F,\mathcal{X}^{cam}_{\hat{a}},\mathcal{X}^{cam}_{a})$ is the generative adversarial loss between $D_{F}$ and $F$. $V_{ cyc}(G,F)$ is the consistency loss to force $F(G(x_a)) \approx x_a$ and $G(F(x_{\hat{a}})) \approx x_{\hat{a}}$.

After the training process for adversarial learning, we will have $(A-1)^2$ generators between each camera pair. 

\subsubsection{Adapt PLS to Adversarial Generated Samples}
With the enhanced dataset, we update our PLS process in three aspects: (1) we make full use of the entire enhanced dataset as training set. (2) more labelled images are available during the initial training process. (3) instead of using the one-shot image feature as sample mining reference, we use the feature centre of that person under different cameras. 

As each image generates images for all $A$ cameras, the size of the final enhanced dataset $\mathcal{\hat{X}}$ is $A$ times the size of the original $\mathcal{X}$. Similarly, we define the new \emph{labelled} set to be $\mathcal{\hat{L}}$ with size $AC$ where $\mathcal{\hat{L}}=\{\mathcal{L},\mathcal{L}^g\}$, $\mathcal{L}^g$(${l}^g \in \mathcal{L}^g$) are the generated images from $\mathcal{L}$. The augmented \emph{unlabelled} set becomes $\mathcal{\hat{U}}$ ($\hat{u} \in \mathcal{\hat{U}}$).

When we use the enhanced dataset during our PLS process, we firstly change the original training set from $\mathcal{X}$ to $\mathcal{\hat{X}}$, with distance matrix $M$ defined in Eq.~\ref{eq:distance} changed to $M^{cam} \in \mathbb{R}^{C \times{(AN-AC)}}$. Each feature distance value $M^{cam}[i,j]$ represents the distance between $l^{ct}_i$ and $\hat{u}_j$ ($\hat{u}_j \in \mathcal{\hat{U}} $), where
\begin{equation}
l^{ct}_i = \frac{1}{A}(\sum_{a=1}^{A-1}{l}^g_{i_a}+{l}_{i})
\end{equation}
is the feature center of the $i$th person, considering all $A$ different cameras. ${l}^g_{i_a}$ is the feature of the generated image for the $a$th camera from ${l}_{i}$.

For the pseudo dataset selection process, for iteration $T$, we select $ATC$ unlabeled samples from $\mathcal{\hat{U}}$ to generate the preliminary pseudo labelled dataset $\mathcal{\hat{U}}^{p}$. The selection rule is similar to vanilla PLS, but we select $AT$ closest samples for each centre feature $l^{ct}$ in iteration $T$.

We use similar relative distance to refine the samples as defined in  Eq.~\ref{eq:refine} of vanilla PLS, the selected pseudo labelled set is $\mathcal{\hat{U}}^{sp}$. The following model training processes stay unchanged.

\subsection{Training Losses}
\label{loss}
The objective of the Re-ID task is different from the traditional classification task. 
In Re-ID, the main purpose is to train a feature extractor, and use that feature to find the same person among a huge number of gallery images. To overcome this problem, several losses have been designed in the past. One  effective loss is MSMLoss \cite{alex2017defense}, which is one type of triplet loss. But MSMLoss is designed for fully-supervised learning instead of semi-supervised learning. In our framework, we use both the softmax loss and our newly designed HSoften-Triplet-Loss for the one-shot Re-ID task. 

A new batch formation rule is designed by taking different nature of labelled samples and pseudo labelled samples into account. In the training process, for each iteration, we randomly sample $B$ \emph{labeled} samples from $\mathcal{L}$. For each \emph{labeled} image, we also randomly select $S-1$ ($(S-1)\leq T$) pseudo labeled samples from $\mathcal{U}^{sp}$ (or from $\mathcal{\hat{U}}^{sp}$ with adversarial learning) in the same class. 

Then, the training batch $\mathcal{X}_{B_S}$ ($x_{i_j} \in \mathcal{X}_{B_S}$) is the combination with both \emph{labelled} images and \emph{pseudo labelled} images. $x_{i_j}$ is a sample with identity index $i$ $ (i\leq B)$ and image index $j$ $(j \leq S)$ in the batch.

\subsubsection{Softmax Loss}

The softmax loss is formulated as:  
\begin{equation}
L_{\emph{\text{softmax}}} = -\sum_{i=1}^{B}\sum_{j=1}^{S}\frac{e^{f(W_{y_{i_j}};\phi (\theta;x_{i_j}))}}{ \sum_{k=1}^{C}e^{f(W_{k};\phi(\theta;x_{i_j}))}}
\label{eq:softmax},
\end{equation} 
where $y_{i_j}$ is the class label of $x_{i_j}$, $W_{k}$ is the weight for class $k$ in the last fully connected layer. 
$f(W_{k};\phi (\theta;x))$ is a combined operation of the batch normalization \cite{Ioffe:2015:BNA:3045118.3045167}, dropout \cite{JMLR:v15:srivastava14a} and fully connected layer (with parameter $W_k$), on top of the feature representation $\phi (\theta;x)$.
\begin{equation}
f(W_{k};\phi (\theta;x)) = W_{k}^T\cdot Dropout_{\gamma}(BN (\phi(\theta;x))
\label{eq:f},
\end{equation} 
where $r$ is the drop rate of the dropout layer.
It is worth to note that in the inference process, we only use $\phi (\theta;x)$ as feature representation.

\begin{figure}[H]
\begin{center}
  \includegraphics[width=0.8\linewidth,trim=300 850 1100 1000,clip]{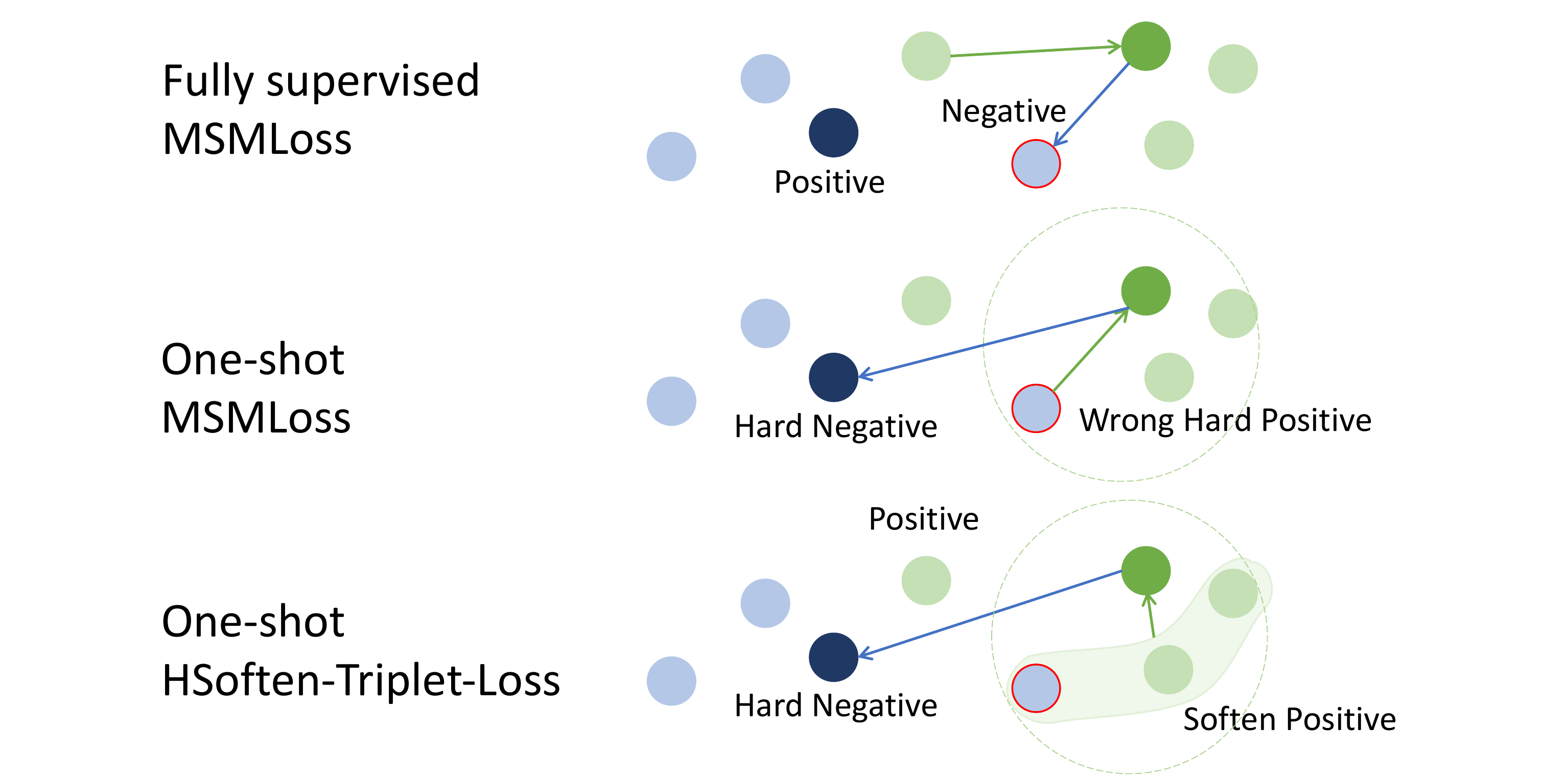}
\end{center}
   \caption{The comparison of different losses.1) In fully-supervised learning,  MSMLoss is perfect to distinct the positive and negative samples. 2) In one-shot learning, an incorrect hard positive sample causes strong miss match. 3) In one-shot learning, soften hard positive can avoid the fatal impact of the incorrect hard positive sample by averaging the features. Best viewed in color.
   }
\label{fig:loss}
\end{figure}

\subsubsection{HSoften-Triplet-Loss}

The MSMLoss \cite{alex2017defense} is formulated as: 
\begin{equation}
\begin{split}
L_{\emph{\text{MSMLoss}}} = \sum_{i=1}^{B}\sum_{j=1}^{S}[\overbrace{\max_{m=1 \dots S}(\left \| \phi(\theta;{x_{i_j})}-\phi(\theta;{x_{i_m})} \right \|_{2}) }^{\emph{\text{hardest positive pair}}} - \\
\overbrace{\min_{n=1 \dots B (n\neq i) \atop m=1 \dots S }(\left \| \phi(\theta;{x_{i_j}})-\phi(\theta;{x_{n_m}}) \right \|_{2}) }^{\emph{\text{hardest negative pair}}} + \alpha],
\label{MSMLoss}
\end{split}
\end{equation} 
which combines both the hardest positive sample selection and the hardest negative sample selection. For each training image $x_{i_j}$ with the ground-truth label, we select the farthest image with the same ID,  $x_{i_m}$, as the hardest positive, and the nearest image with different ID, $x_{n_m}$ ($n\neq i$), in this batch as the hardest negative. The $\left \| \cdot \right \|_{2}$ is the Euclidean distance of two features for the images we selected. $\alpha$ is the hyper-parameter of the margin loss.

HSoften-Triplet-Loss is newly designed for our one-shot learning, which is based on MSMLoss. During the training iterations, the label accuracy of \emph{pseudo labelled} set $\mathcal{U}^{sp}$ (or $\mathcal{\hat{U}}^{sp}$) will gradually drop. Thus, the selected hardest positive $x_{i_m}$ could be with incorrect label, which introduces huge noise. To solve this problem, we design a soft version of hard positive sample feature representation:
\begin{equation}
\hat{\phi}(\theta;{x_{i}}) = \frac{1}{S}\sum_{j=1}^{S}\phi(\theta;{x_{i_j}})
\label{hsoft},
\end{equation}
where $\hat{\phi}(\theta;{x_{i}})$ is the average feature of all samples with identity $i$ in this batch. The final HSoften-Triplet-Loss is:
\begin{equation}
\begin{split}
L_{\emph{\text{HSoft}}} = \sum_{i=1}^{B}\sum_{j=1}^{S}[\overbrace{(\left \| \phi(\theta;{x_{i_j}})-\hat{\phi}(\theta;{x_{i}}) \right \|_{2}) }^{\emph{\text{soften positive pair}}}\\ - 
\overbrace{\min_{n=1 \dots B (n\neq i) \atop m=1 \dots S }(\left \| \phi(\theta;{x_{i_j}})-\phi(\theta;{x_{n_m}})\right \|_{2}) }^{\emph{\text{hardest negative pair}}} + \alpha].
\end{split}
\label{eq:HSoften-triplet}
\end{equation} 

\textbf{The overall loss} is the combination of both softmax and our HSoften-Triplet-Loss.
\begin{equation}
L_{\emph{\text{Overall}}} = L_{\emph{\text{Softmax}}} + L_{\emph{\text{HSoft}}} 
\label{eq:loss}.
\end{equation}

\section{Experiments}
\subsection{Experiment Settings}
\label{Setting}
Our method is evaluated on the test sets of two widely-used datasets for person Re-ID, including Market1501 and DukeMTMC-ReID. \textbf{Market-1501} \cite{zheng2015scalable} contains 32,668 images with 1,501 person IDs, where 751 person IDs are used as the training set, and 750 person IDs are used as the testing set. All pictures are captured from 6 cameras, and each person is annotated in the form of bounding box with the size of 256*128. The number of images with the same person ID varies from 5 to 40, and each person ID contains 7.2 images on average. \textbf{DukeMTMC-reID} \cite{zheng2017unlabeled} is the sub-dataset of DukeMTMC \cite{ristani2016MTMC}, including 36,411 images with 1,401 person IDs, where 702 person IDs are used as the training set, and 699 person IDs are used as the testing set. Different from Market1501, all pictures in DukeMTMC-ReID are captured from 8 cameras, and all images are in different sizes. Moreover, the image number variation for different IDs is much larger, which makes it a more complex and difficult dataset. 

We use both the Cumulative Matching Characteristic (CMC) curve and the mean Average Precision (mAP) to evaluate the performance. The CMC curve is the precision of correct matching with different ranking numbers. Normally, the Re-ID task chooses Rank-1, Rank-5 and Rank-10 to represent CMC curve. The mAP is the mean of the average precision (AP) for all query images. 

\subsection{Implementation Details}
\label{Implementation}
The backbone of our Re-ID model is a ResNet-50 model pre-trained on the ImageNet dataset \cite{imagenet_cvpr09}. The original last fully connected layer of ResNet-50 model is replaced by a new fully connected layer with dropout and batch normalization according to Eq.~\ref{eq:f}, and the paramaters of the new fully connected layer is $2048\times751$ for Market-1501 dataset or $2048 \times 702$ for DukeMTMC-reID dataset due to different number of classes. The dropout rate is 0.5. 

The optimizer is Adam with the momentum of 0.7. The learning rate of the backbone ResNet-50 model is 0.00035, and the learning rate to train the new fully connected layer is 0.0035. For the HSoften-Triplet-Loss, the batch size $B$ is 16 with the number of selected pseudo labelled samples being $S=6$. The margin hyper-parameter $\alpha$ is 0.3.

As the original dataset does not provide an official one-shot selection strategy, we randomly choose the one-shot examples for each class without considering the size of the picture or the camera ID. The one-shot dataset will be available yo reproduce our result.

\subsection{Comparisions} 
\label{results}

\subsubsection{Comparison with State-of-the-Arts} 
\label{overall_results}

In this section, we compare our method with other state-of-the-art person Re-ID methods, which are classified into two groups. The methods in the first group are trained relying on the one-example learning strategy including EUG \cite{wu2019progressive}, while the methods in the second group are trained relying on the transfer learning strategy, which requires much more labels for training than one-example learning strategy, including PUL \cite{Fan_2018}, SPGAN \cite{deng_2018}, TJ-AIDL \cite{wang_2018} and ECN \cite{zhong2019invariance}.

As can observed from Table \ref{tab:one}, among the methods in the lower group (one-shot learning), our model achieves a new state-of-the-art performance on both Market1501 (mAP of 42.7\%) and DukeMTMC-ReID (mAP of 40.3\%). Compared with the previous state-of-the-art method EUG \cite{wu2019progressive}, our method improves the accuracy of mAP by 16.5 on Market1501, and by 11.8 on DukeMTMC-ReID, which shows the robustness of our method on different testing datasets. In terms of the comparison in the second group, our method also achieves competitive results. On both dataset, our method virtually achieves the same accuracy as the best performance method in the upper group (transfer learning), while our method needs much fewer labels for training, which demonstrates the data efficiency of our method.  

\begin{table}[H]
  \caption{Comparison with the state-of-the-art methods on two main datasets Market1501 and DukeMTMC-ReID}
  \label{tab:one}
    \scalebox{0.7}{
    \begin{tabular}{c|c|c|c|c|c|c|c|c|c}
    \hline
    \multirow{2}{*}{Method}&\multirow{2}{*}{Labels}&
    \multicolumn{4}{c|}{Market1501}&\multicolumn{4}{c}{ DukeMTMC-ReID}\cr
    \cline{3-10}
    &&rank-1&rank-5&rank-10&mAP&rank-1&rank-5&rank-10&mAP\cr
    \hline
    \hline
    PUL \cite{Fan_2018}&Transfer&44.7&59.1&65.6&20.1&30.4&46.4&50.7&16.4\cr
    SPGAN \cite{deng_2018}&Transfer&51.5&70.1&76.8&22.8&41.1&56.6&63.0&22.3\cr
    TJ-AIDL \cite{wang_2018} &Transfer&56.7&75.0&81.8&27.4&45.3&59.8&66.3&24.7\cr
    ECN \cite{zhong2019invariance}&Transfer&{\bf 75.1}&{\bf87.6}&{\bf91.6}&{\bf43.0}&{\bf63.3}&{\bf75.8}&{\bf80.4}&{\bf40.4}\cr
    \hline
    EUG \cite{wu2019progressive}&One-shot&55.8&72.3&78.4&26.2&48.8&63.4&68.4&28.5\cr
    {\bf Ours}&One-shot&{\bf 74.6}&{\bf 86.3}&{\bf 90.1}&{\bf 42.7}&{\bf 64.6}&{\bf 75.2}&{\bf 79.1}&{\bf 40.3}\cr
    \hline
  \end{tabular}
  }
\end{table}

\subsubsection{Ablation Study on Components} 
\label{individual_results}

We conduct this ablation study on both Market1501 and DukeMTMC-ReID datasets where parts of our method are disabled or replaced by inferior counterparts to investigate the impact of each component in our method. In detail, after the pseudo label sampling (PLS), we train the model on the normal softmax loss and MSMLoss loss because the PLS is specifically designed for triplet based selection, obtaining the method \emph{PLS+MSMLoss}. Then, in order to adjust to the one-shot Re-ID task, we replace the MSMLoss by the proposed HSoften-Triplet-Loss, obtaining the method \emph{PLS+HSoften-Triplet-Loss}. Finally, we augment the original training set utilizing CycleGAN, and then train the model relying on the HSoften-Triplet-Loss, obtaining the method \emph{PLS+HSoften-Triplet-Loss+CycleGAN}. 
Our \emph{full-labelled baseline} is the fully-supervised learning baseline with all images labeled, using both softmax loss and MSMLoss. The \emph{one-shot baseline} is to directly train the model with only the one-shot labelled images. 
Compared with \emph{one-shot baseline}, our method \emph{PLS+MSMLoss} improves mAP by 34.8 on Market1501 dataset and by 34.0 on DukeMTMC-ReID dataset, which proves the significant contribution of our pseudo labelled sample mining process designed for the triplet. Its mAP, however, is still lower than method \emph{PLS+HSoften-Triplet-Loss} by 1.7 on Market1501 and by 1.5 on DukeMTMC-ReID dataset, which demonstrates that our HSoften-Triplet-Loss is more suitable for the one-shot Re-ID task than the MSMLoss. By incorporating the training set augmentation using CycleGAN, our final method (\emph{PLS+HSoften-Triplet-Loss+CycleGAN}) achieves the mAP of 42.7 and 40.3 on Market1501 dataset and DukeMTMC-ReID, receptively, with a huge improvement against the previous state-of-the-art method, EUG \cite{wu2019progressive}.

\begin{table}[H]
  \centering
  \caption{Ablation study results on Market1501 and DukeMTMC-ReID.}
  \label{tab:method}
    \scalebox{0.6}{
   \begin{tabular}{c|c|c|c|c|c|c|c|c}
    \hline
    \multirow{2}{*}{Method}&
    \multicolumn{4}{c|}{Market1501}&\multicolumn{4}{c}{ DukeMTMC-ReID}\cr
    \cline{2-9}
    &rank-1&rank-5&rank-10&mAP&rank-1&rank-5&rank-10&mAP\cr
    \hline
    \hline
    Full-labelled baseline &93.1&97.7&98.5&81.5&85.2&93.3&95.3&71.2\cr
    One-shot baseline &9.8&20.2&27.6&3.8&8.5&16.7&21.1&3.7\cr \hline
    EUG \cite{wu2019progressive}&55.8&72.3&78.4&26.2&48.8&63.4&68.4&28.5\cr 
    PLS + MSMLoss &68.4&82.5&87.0&38.6&61.0&72.4&76.9&37.7\cr
    PLS + HSoften-Triplet-Loss &69.1&82.7&86.9&40.3&62.2&73.2&78.1&39.2\cr
    {\bf PLS+HSoften-Triplet-Loss+CycleGAN}&{\bf 74.6}&{\bf 86.4}&{\bf 90.1}&{\bf 42.7}&{\bf 64.6}&{\bf 75.2}&{\bf 79.1}&{\bf 40.3}\cr
    \hline
  \end{tabular}
  }
\end{table}

\begin{figure}[H]
\centering
\ref{named}
\captionsetup[subfigure]{font=footnotesize}
\subcaptionbox{}[.4\textwidth]{%
\begin{tikzpicture}[scale=.5]
\begin{axis}[
    xlabel={Ratio [\%]},
    ylabel={Rank 1},
    xmin=0, xmax=100,
    ymin=20, ymax=80,
    xtick={0,20,40,60,80,100},
    ytick={20,30,40,50,60,70,80},
    ymajorgrids=true,
    grid style=dashed,
    legend columns=-1,
    legend entries={PLS+HSoften-Triplet-Loss+CycleGAN;,PLS + HSoften-Triplet-Loss;,PLS + MSMLoss;,EUG \cite{wu2019progressive}},
    legend to name=named,
    legend style={draw=none},
    legend style={draw=none,nodes={scale=0.5}},
]

\addplot[
    color={rgb,255:red,50;green,177;blue,101},
    mark=*,
    ]
    coordinates {
    (0,30.6)(10,57.2)(20,62.3)(30,65.2)(40,67.5)(50,68.3)(60,69.3)(70,70.3)(80,72)(90,74.6)
    };
\addplot[
    color={rgb,255:red,56;green,167;blue,208},
    mark=triangle,
    ]
    coordinates {
    (0,31.5)(10,46.7)(20,54.7)(30,58.7)(40,59.0)(50,60.8)(60,63.3)(70,65.3)(80,66.4)(90,68.4)
    };
\addplot[
    color={rgb,255:red,246;green,112;blue,136},
    mark=square,
    ]
    coordinates {
    (0,29.5)(10,49.0)(20,54.8)(30,58.7)(40,59.1)(50,60.7)(60,61.9)(70,63.5)(80,65.6)(90,66.4)
    };
\addplot[
    color={rgb,255:red,206;green,143;blue,49},
    mark=star,
    ]
    coordinates {
    (0,25)(10,37)(20,44.8)(30,47.5)(40,50.3)(50,50)(60,51)(70,50.8)(80,50)(90,46)
    };
\end{axis}
\end{tikzpicture}}
\hspace{.2in}
\subcaptionbox{}[.4\textwidth]{
\begin{tikzpicture}[scale=.5]
\begin{axis}[
    xlabel={Ratio [\%]},
    ylabel={mAP [\%]},
    xmin=0, xmax=100,
    ymin=0, ymax=50,
    xtick={0,20,40,60,80,100},
    ytick={0,10,20,30,40,50},
    ymajorgrids=true,
    grid style=dashed,
]
\addplot[
    color={rgb,255:red,50;green,177;blue,101},
    mark=*,
    ]
    coordinates {
    (0,11.9)(10,24.1)(20,28.1)(30,31.2)(40,33.3)(50,34.3)(60,35.5)(70,36.2)(80,39.5)(90,42.7)
    };
\addplot[
    color={rgb,255:red,56;green,167;blue,208},
    mark=triangle,
    ]
    coordinates {
    (0,13.1)(10,20.6)(20,23.2)(30,26.8)(40,29.1)(50,30.9)(60,32.8)(70,34.9)(80,37.2)(90,39.8)
    };
\addplot[
    color={rgb,255:red,246;green,112;blue,136},
    mark=square,
    ]
    coordinates {
    (0,11.8)(10,21.5)(20,26.1)(30,27.3)(40,29.5)(50,30.1)(60,31.7)(70,33.5)(80,35.0)(90,38.3)
    };
\addplot[
    color={rgb,255:red,206;green,143;blue,49},
    mark=star,
    ]
    coordinates {
    (0,9)(10,15)(20,18)(30,21)(40,22)(50,23)(60,24)(70,24)(80,23)(90,22)
    };
\end{axis}
\end{tikzpicture}}
\subcaptionbox{}[.4\textwidth]{
\begin{tikzpicture}[scale=.5]
\begin{axis}[
    xlabel={Ratio [\%]},
    ylabel={Precision},
    xmin=0, xmax=100,
    ymin=20, ymax=100,
    xtick={0,20,40,60,80,100},
    ytick={20,30,40,50,60,70,80,90,100},
    ymajorgrids=true,
    grid style=dashed,
]
\addplot[
    color={rgb,255:red,50;green,177;blue,101},
    mark=*,
    ]
    coordinates {
    (0,97.1)(10,92.3)(20,83.7)(30,75.1)(40,73)(50,69.2)(60,64.9)(70,62.3)(80,60.6)(90,61.6)
    };
\addplot[
    color={rgb,255:red,56;green,167;blue,208},
    mark=triangle,
    ]
    coordinates {
    (0,85.5)(10,80.5)(20,72.2)(30,68.4)(40,64.1)(50,60.5)(60,58.1)(70,55.8)(80,55.5)(90,56.8)
    };
\addplot[
    color={rgb,255:red,246;green,112;blue,136},
    mark=square,
    ]
    coordinates {
    (0,83.8)(10,79.53)(20,71.51)(30,68.56)(40,63.9)(50,60.1)(60,57.4)(70,54.5)(80,53.8)(90,54.8)
    };
\addplot[
    color={rgb,255:red,206;green,143;blue,49},
    mark=star,
    ]
    coordinates {
    (0,70)(10,62)(20,60)(30,58)(40,52)(50,50)(60,48)(70,46)(80,40)(90,37)
    };
\end{axis}
\end{tikzpicture}}
\hspace{.2in}
\subcaptionbox{}[.4\textwidth]{
\begin{tikzpicture}[scale=.5]
\begin{axis}[
    xlabel={Ratio [\%]},
    ylabel={Recall},
    xmin=0, xmax=100,
    ymin=0, ymax=70,
    xtick={0,20,40,60,80,100},
    ytick={0,10,20,30,40,50,60,70},
    ymajorgrids=true,
    grid style=dashed,
]
\addplot[
    color={rgb,255:red,50;green,177;blue,101},
    mark=*,
    ]
    coordinates {
    (0,4.6)(10,8.7)(20,16.1)(30,25.1)(40,27.8)(50,32.6)(60,38.7)(70,43.1)(80,48.7)(90,55.1)
    };
\addplot[
    color={rgb,255:red,56;green,167;blue,208},
    mark=triangle,
    ]
    coordinates {
    (0,4.7)(10,8.8)(20,15.9)(30,21.9)(40,24.5)(50,29.4)(60,33.7)(70,38.8)(80,47.3)(90,52.5)
    };
\addplot[
    color={rgb,255:red,246;green,112;blue,136},
    mark=square,
    ]
    coordinates {
    (0,3.8)(10,7.7)(20,15.7)(30,19.8)(40,24.4)(50,31)(60,32.2)(70,35.8)(80,43.4)(90,49.4)
    };
\addplot[
    color={rgb,255:red,206;green,143;blue,49},
    mark=star,
    ]
    coordinates {
    (0,0)(10,7)(20,13)(30,18)(40,24)(50,27)(60,30)(70,33)(80,35)(90,37)
    };
\end{axis}
\end{tikzpicture}}
\caption{Effects of the ratio of pseudo labelled samples over all unlabelled samples on Market1501, our methods are compared with EUG \cite{wu2019progressive}. (a) and (b) are Rank-1 and mAP scores of inference, respectively. (c) and (d) are the precision and recall of the selected pseudo labelled samples, respectively. Best viewed in color.
}
\label{fig:lines}
\end{figure}
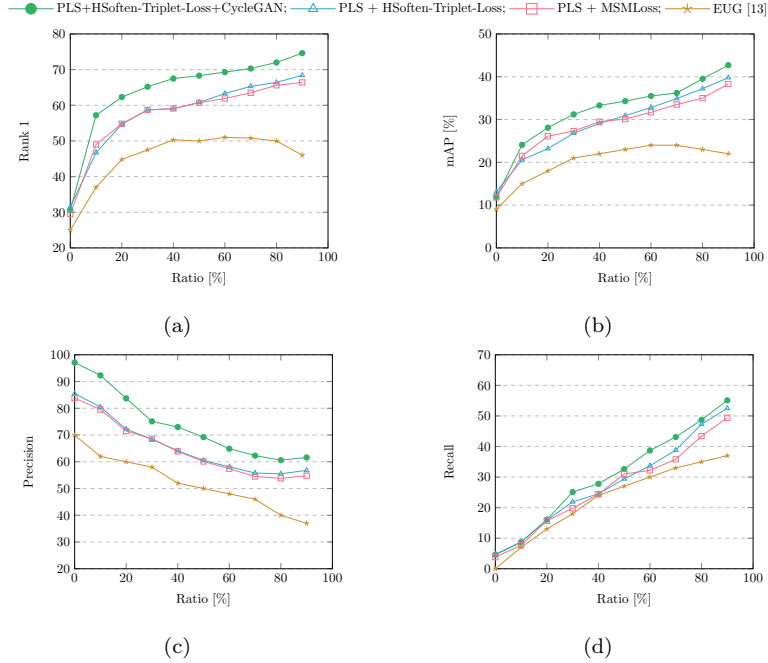

We also conduct experiments to study how the ratio of the pseudo labelled samples over all unlabelled samples affects the final performance. As can be observed from Fig.~\ref{fig:lines}, our methods outperforms EUG \cite{wu2019progressive} significantly with all the pseudo labelled sample ratio under all metrics including  rank-1, mAP, precision and recall. Meanwhile, as illustrated in Fig.~\ref{fig:lines}.(a) and Fig.~\ref{fig:lines}.(b), EUG reaches its top Rank-1 and mAP scores when the pseudo labelled samples ratio is around 70\%, and both scores drop after that. For our methods, both scores constantly increase with the pseudo labelled samples ratio, which proves that our method is stable enough to make full use of all the training data. 

In Fig.~\ref{fig:lines}.(c) and Fig.~\ref{fig:lines}.(d), we study the precision and recall of the selected pseudo labelled samples respectively. Fig.~\ref{fig:lines}.(c) demonstrates that, MSMloss performs similarly as HSoften-Triplet-Loss when the ratio of the pseudo labelled samples over all unlabelled samples is below 60\%, because the pseudo labels are with high accuracy at the beginning. But HSoften-Triplet-Loss outperforms MSMloss when the ratio goes higher, which indicates that HSoften-Triplet-Loss is pretty stable when the number of incorrect pseudo labelled samples increases. In terms of recall, as can be observed from the Fig.~\ref{fig:lines}.(d), the curve of EUG resembles the log function, which greatly constricts its upper border. Our method, however, can continue to achieve a higher recall with a higher pseudo label ratio, leading to a much higher upper border.

\subsubsection{Visualization of Pseudo Labelled Samples} 
\label{Visualization}

\begin{figure}[H]
\begin{center}
  \includegraphics[width=1\linewidth,trim=100 200 700 1200,clip]{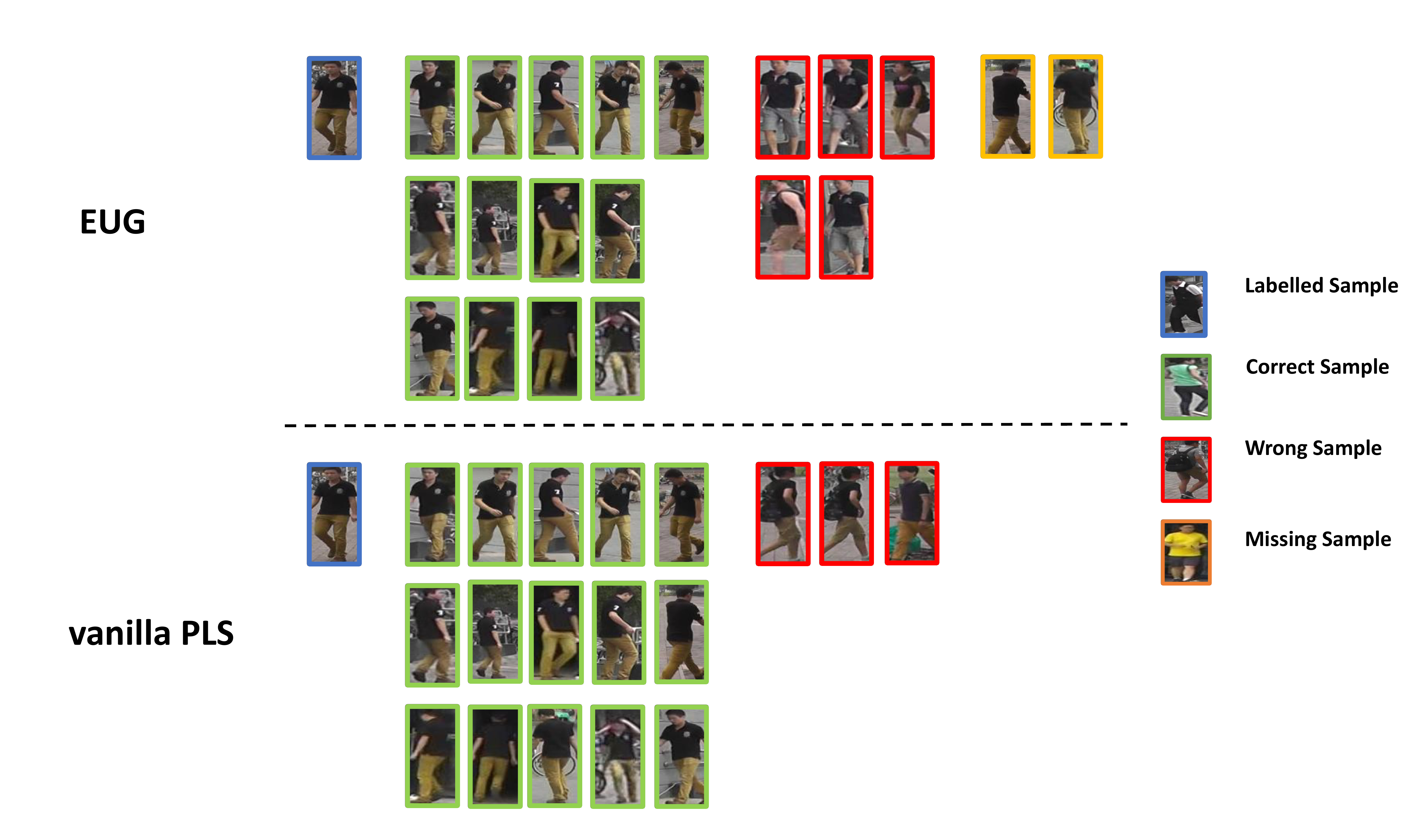}
\end{center}
   \caption{Case study: Comparison between pseudo labelled selections of our vanilla PLS and EUG \cite{wu2019progressive}. Best viewed in color.
   }
\label{fig:compare}
\end{figure}

Fig.~\ref{fig:compare} reports a case study of the pseudo labelled samples. Our method performs much better than EUG \cite{wu2019progressive} in terms of pseudo labelled sample selection. Firstly, our method successfully obtains all the images of a particular person, but EUG misses two of them. Also, the number of wrong samples selected by our method is only around half of EUG. Note that the wrong images selected by our method is much more similar to the reference image with both black T-shirt and brown pants, but EUG chooses the person with grey shorts, which greatly increases the difficulty of the further training process.

\subsubsection{Performance with More Shots} 
\label{more-shots}

To explore the influence of the quantity of the labelled samples in the training set, we evaluate our model using several few-shots settings, i.e. two-shots, three-shots and five-shots. As can be observed from Table \ref{tab:times}, with the additional labelled samples provided by these few-shot settings, the mAP score of our method on Market1501 dataset sees a huge increase by 18.1, which indicates the significence of labelled samples. 

\begin{table}[H]
  \centering
  \caption{Semi-supervised method results with different number of shots.}
  \label{tab:times}
    \begin{tabular}{c|c|c|c|c}
    \hline
    \multirow{2}{*}{Settings}&
    \multicolumn{4}{c}{Market1501}\cr
    \cline{2-5}
    &rank-1&rank-5&rank-10&mAP\cr
    \hline
    One-shot&74.6&86.4&90.1&42.7\cr \hline
    Two-shots&79.5&90.3&93.4&51.5\cr
    Three-shots&80.6&91.8&95.0&56.2\cr
    Five-shots&84.2&92.9&97.1&60.8\cr
    \hline
  \end{tabular}
\end{table}

\section{Conclusion}
In this paper, we propose a novel pseudo label sampling process with a new triplet loss HSoften-Triplet-Loss, which is specifically designed to address the one-shot person re-identification task. It proves that our sampling process is more suitable for triplet selection, and the HSoften-Triplet-Loss is more robust when dealing with the incorrect pseudo labelled samples. In addition, we further adopt an adversarial learning network to provide more samples with different ID for the training set, which increases the diversity of the training set. Our method boosts the performance against the previous state-of-the-art method on mAP by 16.5 on Market1501 dataset, and by 11.8 on DukeMTMC-Reid dataset.

\bibliography{mybib}

\end{document}